\documentclass[]{llncs}
\usepackage{wrapfig}
\usepackage{multirow}
\usepackage{graphicx}
\usepackage{amsfonts}
\usepackage{mathtools}
\usepackage{amsmath}
\usepackage{amssymb}
\usepackage{cite}
\usepackage{array}
\usepackage{subfigure}
\usepackage{setspace}
\usepackage{booktabs}
\usepackage[misc,geometry]{ifsym} 
\usepackage[ruled,vlined]{algorithm2e}
\usepackage{ulem}
\usepackage[table,xcdraw]{xcolor}
\usepackage{caption}

\usepackage[colorlinks,
            linkcolor=green,
            anchorcolor=red,
            citecolor=blue
            ]{hyperref}

\begin{document}
\title{Not just Birds and Cars: Generic, Scalable and Explainable Models for Professional Visual Recognition}

\author{Junde Wu \inst{1,2,3} \and
Jiayuan Zhu \inst{2}  \and 
Min Xu \inst{3} \and
Yueming Jin \inst{1}
}

\authorrunning{J. Wu et al.}

\institute{National University of Singapore \\
\and
University of Oxford\\
\and
MBZUAI \\
}

\maketitle             
\begin{abstract}
Some visual recognition tasks are more challenging then the general ones as they require professional categories of images. The previous efforts, like fine-grained vision classification, primarily introduced models tailored to specific tasks, like identifying bird species or car brands with limited scalability and generalizability. This paper aims to design a scalable and explainable model to solve Professional Visual Recognition tasks from a generic standpoint. We introduce a biologically-inspired structure named Pro-NeXt and reveal that Pro-NeXt exhibits substantial generalizability across diverse professional fields such as fashion, medicine, and art—areas previously considered disparate. Our basic-sized Pro-NeXt-B surpasses all preceding task-specific models across 12 distinct datasets within 5 diverse domains. Furthermore, we find its good scaling property that scaling up Pro-NeXt in depth and width with increasing GFlops can consistently enhances its accuracy. Beyond scalability and adaptability, the intermediate features of Pro-NeXt achieve reliable object detection and segmentation performance without extra training, highlighting its solid explainability. We will release the code to foster further research in this area. 
\end{abstract}

\section{Introduction}
\label{sec:intro}

\begin{figure}[h]
\centering
\includegraphics[scale=0.2]{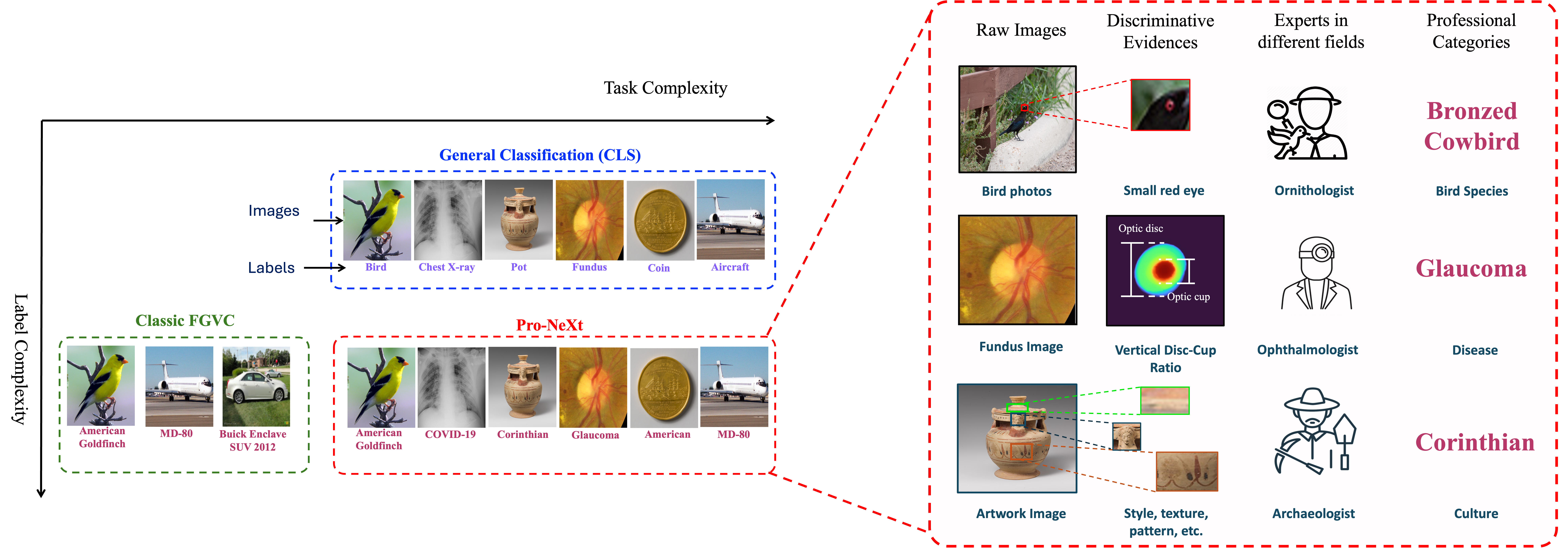}
\caption{ Generic Professional Recognition is more challenging than general and FGVC as it faces both Label Complexity (professional labels) and Task Complexity (a wide range of tasks).}
\label{fig:facial}
\vspace{-15pt}
\end{figure}

Deep neural networks (DNNs) have made significant progress in computer vision tasks. Image classification, as a fundamental task in this field, has been extensively studied by various methods. Thanks to the efforts of these researchers, it has achieved an exceptionally high level, which is comparable to that of humans in some cases~\cite{geirhos2017comparing}. However, some visual classification tasks are challenging even for regular individuals in human society. These tasks can only be completed by a small group of experts who have received extensive professional training. For instance, only ornithologists can accurately identify the species of a wild bird in a photo; only ophthalmologists can diagnose glaucoma from fundus images; and only archaeologists can determine the cultural origin of an artwork from a picture of it. Since these professional categories are semantically more complex than general ones, these tasks pose a greater challenge for deep learning models.

The previous efforts, like Fine-Grained Vision Classification (FGVC), predominantly proposed \textit{task-specific} models confined to a few \textit{homogeneous} benchmarks, such as recognizing the bird species or the vehicle brands. Moreover, most of these methods \cite{zhang2019learning,ge2019weakly,chen2019destruction,du2020fine,luo2019cross} hard to scale-up to consume a large amount of diverse data. As a result, previous FGVC methods show limited generalization over diverse tasks need professional categories. (Experimental supportive results shown in Table \ref{tab:mainres}). As illustrated in the taxonomy in Fig.~\ref{fig:facial}, while current FGVC methods predict more professional labels (higher label complexity) than general classification, they can only handle a limited range of tasks (lower task complexity).

The current research gap motivates us to seek a more universal solution capable of addressing both high task complexity and label complexity. In other words, it can generally handle all tasks involving the inference of a professional category from salient objects in images. Illustrated in the taxonomy in Fig.~\ref{fig:facial}, Generic Professional Recognition is more challenging than both general classification tasks and traditional FGVC, as it needs to handle both task and label complexities.

Through carefully observing the unique features of the Generic Professional Recognition data, we find the challengers can be summarized and exemplified by three typical cases. First and also the obvious one is that the devil is in the details. A small feature in the image can determine the category of the entire image, as in the case of birds shown in Fig.\ref{fig:facial}. Secondly, not only local details but also global structures matter. Such as the case of fundus image shown in Fig.\ref{fig:facial}, a large vertical optic cup and disc ratio (vCDR) is a significant biomarker for glaucoma screening. Lastly, feature-level interactions are also crucial. For instance, as the artwork example in Fig.~\ref{fig:facial}, a comprehensive analysis of various features such as painting style, object textures, and patterns is necessary to identify the Corinthian culture of the pot.

Though these properties seem individual to each other, we find that they can be ascribed to a unifying principle in natural biology, given that human experts only conduct one unique behaviour to discern more professional categories: they call more times of visual hierarchy mechanism \cite{djamasbi2011visual}. Inspired by this observation, we propose Pro-NeXt Model that mimics the visual hierarchy mechanism of human experts in discerning the category of objects. In specific, we design a module named Gaze-Shift, to progressively zoom in on the salient parts and memorize the remaining context in a hierarchical manner. In the end, we fuse the final focal feature and all the memorized context features to make the final decisions. Such a simple and unique design facilitates the Pro-NeXt Model to outperform SOTA with solid scalability and generalization ability.

The contributions of the paper can be concluded as follows:
1) We propose Pro-NeXt Model, which is a generic framework showing the potential to recognize all professional categories from a wide range of domains.
2) Pro-NeXt possesses good scalability: It's generalization ability and task-specific performance consistently increase by simply scaling-up the model.
3) Pro-NeXt possesses good explainability: It's immediate features show strong object detection and segmentation ability without extra training or feature extraction, even outperforms many carefully-designed week-supervision methods.
4) We achieve new SOTA on 12 benchmarks within 5 different domains. Our basic-size model Pro-NeXt-B outperforms all task-specific models.

\section{Related Work}
Fine-grained visual categorization (FGVC) aims to distinguish subordinate categories within entry-level categories. Classic FGVC strategies are mainly based on learning the regional attention (zoom-in), to encourage the network to focus on the crucial details for the professional recognition ~\cite{zhang2019learning,ge2019weakly,chen2019destruction,du2020fine,luo2019cross}. This strategy proves highly effective on main-stream FGVC benchmarks where regional details play a pivotal role, such as CUB~\cite{wah2011caltech}, FGVC-Aircraft~\cite{maji2013fine}, and Stanford-Cars~\cite{krause20133d}. However, such property is not saliently observed on many other professional recognition tasks, like diseases prediction or artifacts attribution, which cause their lose of generalization on such tasks and many others. In these tasks, alignment with the FGVC definition often goes unrecognized. Researchers commonly propose task-specific models without acknowledging any connection to FGVC. For instance, \cite{zhu2022dual, fu2018disc, fan2022detecting, jiangsatformer} introduce specific networks for glaucoma detection, while \cite{zhang2021transformer, ahmed2021convid, tian2018psnet} propose unique models for COVID-19 prediction. 
We believe this stems from a lack of cross-disciplinary understanding and aim to bridge this gap by offering a unified solution for all these tasks that share a similar nature.

Strong empirical evidence that increasing the scale of models or data in deep learning benefits generalization and transfer learning. Initial studies, like \cite{hestness2017deep}, highlighted a power law relationship between scale and model performance. Empirical and theoretical work, particularly on large language models \cite{brown2020language}, revealed broad generalization capabilities with larger scales. Scaling laws were extended to vision models, enabling accurate prediction of performance at larger scales \cite{dehghani2023scaling, peebles2023scalable}. Studies also found that increasing scale improved performance on downstream tasks, though upstream performance doesn't always correlate. However, most previous methods designed for professional recognition are hard to directly scale up. In addition, there is still lack of the specific study of how to reproduce the scalable models for the professional recognition to the best of our knowledge.

\section{Method}
\subsection{Motivation}\label{sec:hier}
When human recognizing an object, vision system will follow a mechanism called "visual hierarchy"~\cite{barghout1999differences,barghout2003global,lindeberg2013computational}. That is we will first shift the attention to a salient object, such as the bird in the picture, but also retain the impression of the context, such as the spring season, in the memory. It shows that when we are doing a more complicated recognition, like an expert is obtaining more professional information from the image, the process will be repeated more times than the ordinary cases
~\cite{barghout2003global}. For example, the expert will further observe the details of birds, such as wings and beaks, and retain the overall impression of birds, such as the blue color, in the memory. Later, more detailed features, such as the patterns of feathers on the wings, will be observed, and so on. Finally, the expert can integrate all the memorized information and the observation to make a comprehensive inference.


\begin{figure*}[h]
\centering
\includegraphics[width=0.85\linewidth]{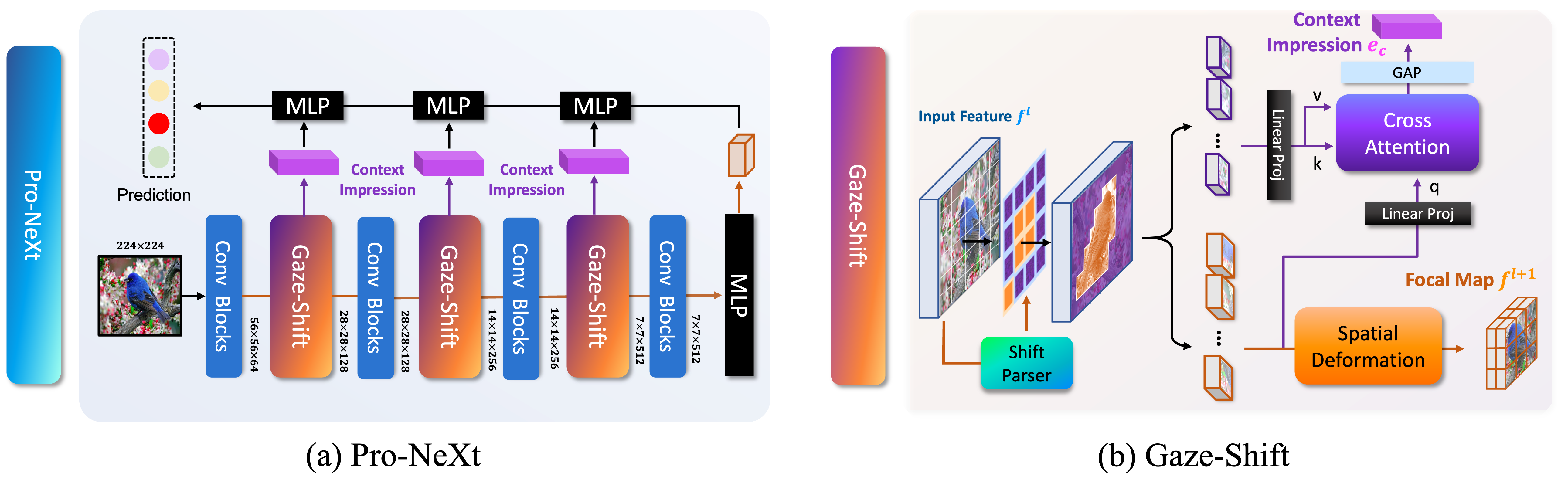}
\caption{
An illustration of our Pro-NeXt framework, which starts from (a) an overview pipeline of Pro-NeXt, and continues with zoomed-in diagrams of (b) Gaze-Shift.}
\vspace{-15pt}
\label{fig:framework}
\end{figure*}


\subsection{Overall Architecture}\label{sec:overall}
Inspired by this biological process, we design Pro-NeXt to hierarchically decouple the focal part and the context information, then individually process the two parts in each stage. This design ensures both the scalability and accuracy of the model on professional recognition, since it not only allows the intermediate context features extracted by scalable transformers directly effect the final prediction, but also zoom-in focal features extracted by CNN pay attention to the details.

An illustration is shown in Fig.~\ref{fig:framework} (a). Over a standard CNN backbone~\cite{he2016deep}, we propose a module named Gaze-Shift between two Convolution Blocks to decompose the $l$ stage feature map $f^{l}$ to the focal-part feature map $f^{l+1}$, called Focal-Map and a context-related embedding $e_{c}^{l}$, named Context-Impression. The Focal-Map will be sent to the next stage and Context-Impression will be memorized. Repeating this way, the local but discriminative parts will be progressively abstracted to high-level semantic features, while the Context-Impression will be stripped out at its appropriate level to reinforce/calibrate the final decision.

Formally, let $\oslash$ denotes the binary masking, and consider Conv (Convolution-like layers), SP (Shift-Parser \ref{sec:nerf}), MLP and CA (Cross-Attention) are parameterized by $\theta_{cv}$, $\theta_{sp}$, $\theta_{mlp}$ and $\theta_{ca}$, respectively. our pipeline can be represented as:
\vspace{-6pt}
\begin{equation}
    f^{l+1} = Conv [(1 - SP(f^{l})) \oslash f^{l} ] ; f^{1} = Conv (x)
\end{equation}
\vspace{-17pt}
\begin{equation}
    e^{l}_{c} = CA [ SP(f^{l}) \oslash f^{l},  (1 - SP(f^{l})) \oslash f^{l}]
\end{equation}
\vspace{-17pt}
\begin{equation}
    e^{pred} = MLP [f^{3} + \sum_{l = 1}^{3} e^{l}_{c}]
\end{equation}
We represent all parameters as $\theta = \{\theta_{cv}, \theta_{sp}, \theta_{mlp}, \theta_{ca}\}$, and update them in an end-to-end manner through the cross-entropy loss function $\mathcal{L}_{ce}$ with one-hot ground truth $e^{gt}$:
\vspace{-6pt}
\begin{equation}
  \theta^{new} = \theta^{old} + \alpha \nabla_{\theta} \mathcal{L}_{ce} (e^{pred}, e^{gt})
\vspace{-6pt}
\end{equation}

\subsection{Gaze-Shift}\label{sec:hier}
The proposed Gaze-Shift works as which shown in Fig. \ref{fig:framework} (b). Given a feature $f\in \mathbb{R}^{H \times W \times C}$, we first packetized the feature map with a patch size $k$. Then we propose a Shift-Parser (Section \ref{sec:nerf}) to generate a binary patch-level map $m \in \mathbb{R}^{p \times p \times 1}$, where $p = H / k$ considering the common case $H$ equals to $W$. Mask $m$ spatially splits the patches of the feature map to focal patches and context patches. In the focus branch, the selected focal patches are spatially deformed (Section \ref{sec:so}) to keep the spatial correlations. In this process, they will be downsampled and abstracted to the focal-part feature $f^{l+1} \in \mathbb{R}^{\frac{H}{2} \times \frac{W}{2} \times 2C}$. In the context branch, the focus patches and context patches are interacted through Cross Attention (CA)~\cite{chen2021crossvit} to obtain $e_{c}$. The focal parts are used as the \textit{query}, the context parts are used as the \textit{key} and \textit{value} to be interacted by the attentive mechanism (see Section \ref{sec:ca} for details). In this way, we encode the discriminative embedding from the context content considering its interaction with the focus parts.


\begin{figure}
\centering
\includegraphics[scale=0.35]{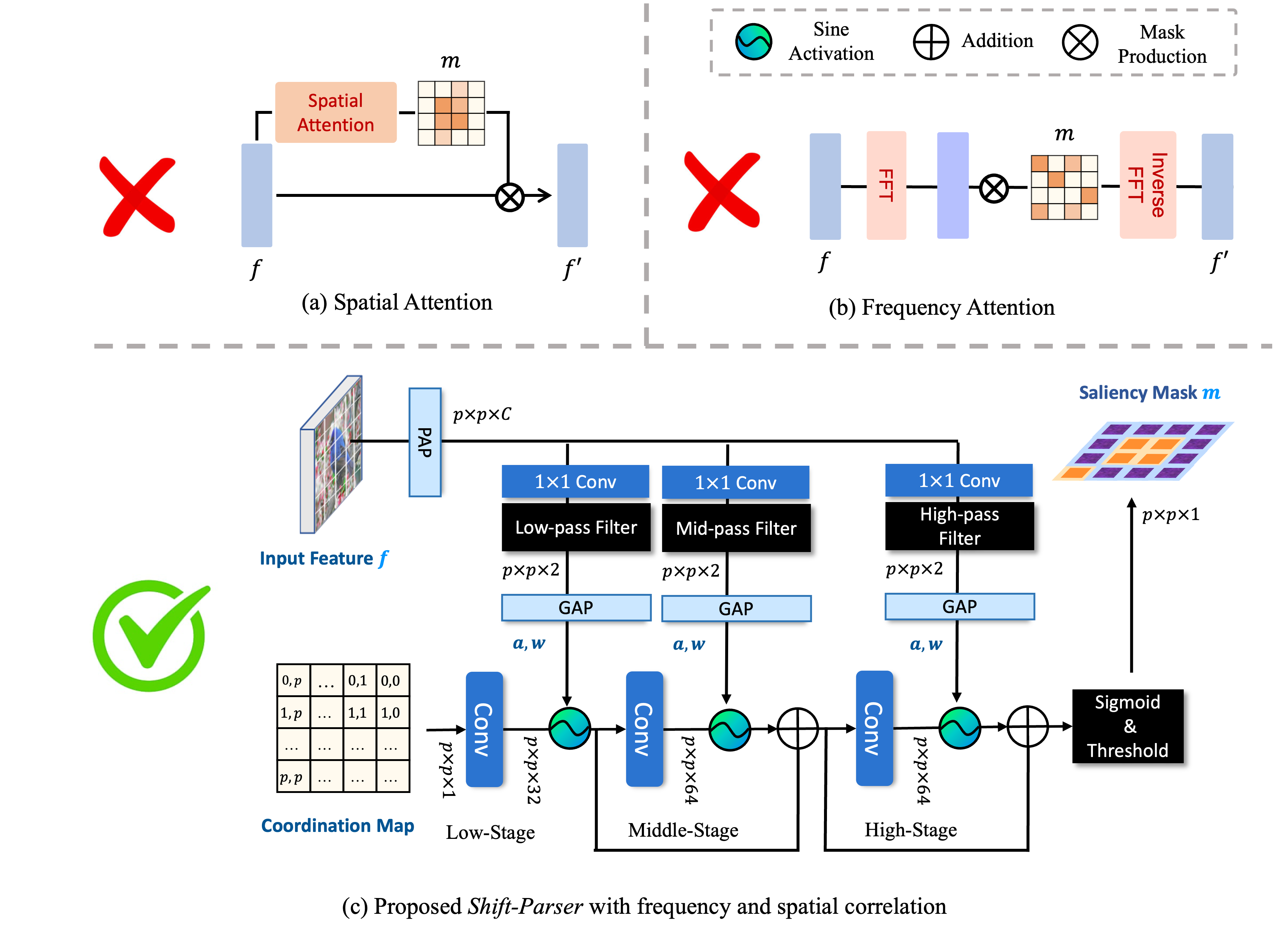}
\caption{A comparison of different feature attentive mechanisms. The proposed Shift-Parser 
(c) takes advantages of both spatial domain methods (a) and frequency domain method (b), which groups the features in frequency domain while also maintaining the spatial constraint.}
\label{fig:freqnerf}
\vspace{-15pt}
\end{figure}


\subsection{Shift-Parser}\label{sec:nerf}
In Gaze-Shift, we employ a learnable parser to distinguish between the part and context features, which we call the Shift-Parser. We establish a binary parser to aware both frequency and spatial correlation since either current spatial and frequency-based modeling methods have their drawbacks. On the one hand, spatial modeling methods \cite{woo2018cbam,liu2022dynamic}) are likely to cause overfitting as the feature maps and attention maps are strongly correlated, as shown in Fig.~\ref{fig:freqnerf} (a). In contrast, processing the features in the frequency space has been shown to be an efficient way to improve generalization \cite{rao2021global,lee2021fnet,tang2022image} recently. However, directly learning the map in the frequency space \cite{rao2021global} would sabotage the spatial correlation of the parser and affect the overall performance, as the focal parts or context parts are likely to cluster together spatially, as shown in Fig.~\ref{fig:freqnerf} (b).

To introduce spatial constraints in the frequency space, we find Neural Fields (NeRF)\cite{xie2022neural} could be a fitting technique to be adopted here. NeRF is a network that projects a coordinate map to a target value map. The inductive bias of the neural networks will force it to produce similar outputs given similar inputs, which naturally constrains the output to be spatially smooth\cite{xie2022neural}. In addition, introducing the frequency knowledge into it is also feasible. We achieve that by designing a conditional NeRF, which is conditioned by the frequency encoding of the given feature map. More specifically, we follow SIREN~\cite{sitzmann2020implicit} to insert a periodic activation function between each two layers, but we control the activation through conditional embedding instead of direct learning. In this way, the periodic activation, which naturally represents the frequency of the output, will be directly affected by the given feature map.

A detailed illustration of the Shift-Parser is shown in Fig.~\ref{fig:freqnerf} (c). It generates a binary patch-level map $m$ conditioned on the input feature map $f$. The map assigns each patch of the feature map to either 'focus' or 'context'. The main architecture for the map generation is a 2D Neural Fields (NeRF) network, which predicts the value of the position from the inputted coordinate map. Unlike the original implementation of NeRF, we apply the convolution layers over the whole coordinate map for efficiency. We include a total of six convolution layers, with one in the low-stage, four in the middle-stage, and one in the high-stage. To introduce the frequency of given feature $f$, we encode $f$ as the parameters to control the amplitude $a$ and frequency $w$ of the sine activation function $a sin(wx)$ between two convolution layers, where $x$ is the forward feature element. To be specific, we first apply patch-average pooling to convert $f$ to a $p\times p\times C$ map, followed by encoding it using $1 \times 1$ convolution to produce a $p \times p \times 2$ map. The two channels represent the amplitude and frequency for each patch. We then control the bandwidth of each stage to encourage coarse-to-fine generation, following the physical concept of band-pass filters. We eliminate the top 20\% high-frequency activations in the low-pass filter, the top 10\% high-frequency and low-frequency activations in the middle-pass, and the top 20\% low-frequency activations in the high-pass filter. We then use global average pooling (GAP)\cite{lin2013network} to produce two values $a$ and $w$ that determine the activation. We also adopt skip connections over the middle and high stages to pass low-frequency information to higher levels. Finally, we use the Sigmoid function and 0.5 thresholding on the last layer to produce a binary $p \times p$ map. The Shift-Parser is trained end-to-end with the whole model supervised by classification labels.

\vspace{-7pt}
\subsubsection{Partial \& Context Feature Processing}\label{sec:details}
\hfill\\
\textbf{Focal Feature with Deformable Convolution}\label{sec:so}
After the parser selects the focal patches, we reform and abstract them to the next stage feature map while maintaining their original spatial correlation. The process is implemented by simple padding, max pooling, and deformable convolution~\cite{dai2017deformable}. 
Specifically, we first arrange the selected focal patches according to their previous positions and adopt padding following 2-stride max pooling to downsample the feature map to $\frac{H}{2}\times \frac{H}{2} \times C$.

Then, we apply a deformable convolution on the feature map to keep the same scale but double the channels. Deformable convolution learns the offsets that help it abstract only the informative positions and ignore the blank ones. In this way, it actually enlarges the selected focus features to the blank positions. More details and experiments about the spatial organization are provided in supplementary material.

\noindent
\textbf{Context Feature with Cross Attention}\label{sec:ca}
We use Cross Attention (CA)~\cite{chen2021crossvit} to model the interaction between the focus and context features in each stage. 
Different from its original setting, we do not use the class token for the classification. Instead, only embeddings (flattened from the feature maps) are interacted through the attention mechanism. GAP is applied to the produced results to obtain the final embedding. 
In addition, we use conditional position encoding~\cite{chu2021conditional} to encode the position information. The flattened focus embedding and flattened context embedding will be concatenated as the condition for positional encoding learning. The produced encoding is split into the focus and context parts based on the positions, and respectively added to the focus embedding and context embedding.

\section{Experiments}
\subsection{Dataset}
We conduct the experiments on 12 professional recognition dataset with 5 different domains, including wild life species classification, vehicle brand classification, disease screening from the medical images, art attribute recognition, and fashion style recognition. For wild life species classification, 4 benchmarks, CUB-200-2011 (CUB)~\cite{wah2011caltech}, iNat2017~\cite{van2018inaturalist}, Kaggle Cassava Leaf (Cas-Leaf), and Stanford-dogs (St-Dogs)~\cite{khosla2011novel} are involved in demonstrating the performance of our method. For vehicle brand classification, FGVC-Aircraft (Air)~\cite{maji2013fine} and Stanford-Cars (St-Cars)~\cite{krause20133d} are involved. In disease screening, we use Covidx dataset~\cite{wang2020covid}, which is a large-scale and open access benchmark dataset for predicting COVID-19 positive cases from chest X-Ray images, and REFUGE2 dataset~\cite{fang2022refuge2}, which is a publicly released challenge dataset for screening glaucoma from fundus images. WikiArt~\cite{saleh2015large} and iMet~\cite{zhang2019imet} are used as two art-related datasets. WikiArt consists of paintings and iMet mainly consists of artworks in The Metropolitan Museum like sculptures or porcelains. For fashion style recognition, we use FashionStyle14\cite{takagi2017makes} and DeepFasion2 \cite{ge2019deepfashion2} for the validation.

\subsection{Experimental Setup}
\subsubsection{Training}
We preprocess images by resizing to $512 \times 512$ pixels and cropping to $448 \times 448$ for model input. The final linear layer initializes with zeros, while other weights follow the ViT standard. Training employs random cropping, and testing uses center cropping. AdamW with specific parameters optimizes the models, and we train with a batch size of 128, using only MixUp and horizontal flips for augmentation. Data imbalance correction was unnecessary for Pro-NeXt's stable training across configurations. Pretraining occurs on ImageNet, followed by retraining on all 12 dataset training sets. The samples are randomly picked in the training process. We find such unified training improves the Pro-NeXt performance although each benchmark has no overlap and the objects are quite different.

\subsubsection{Configuration}
We investigate the hyper-parameters within the Pro-NeXt architecture and study the scaling properties of our model variants. We name the models according to their size and latent patch sizes $p$, e.g., Pro-NeXt-L/4 refers to the larger model with p = 4.
We build our basic model the Context Impression with $3$ CA layers, each with $12$ heads and $784$ embedding length. We scaling the cross-attention and deformable convolution in the Gaze-Shift to get a series of models with different sizes. The models are denoted as $S$, $B$, $L$, $XL$, and $H$ for the increasing sizes. Detailed model setting is in our supplementary.

\subsubsection{Compute}
We implement all models on PyTorch platform and trained models by NVIDIA A100 GPUs. Our most compute intensive model, Pro-NeXt-H is trained on 11 nodes, each with 8 GPUs.


\begin{table*}[!t]
\caption{The comparison of Pro-NeXt with SOTA classification methods in different domains. The gray background denotes the task-specific methods. The number of the parameters is counted in million (M).}\label{tab:mainres}
\centering
\resizebox{0.95\textwidth}{!}{
\begin{tabular}{c|c|c|cccc|cc|cc|cc|cc|c}
\toprule
\hline
\multicolumn{1}{c|}{Method}     & Param       & Architecture               & \multicolumn{4}{c|}{Wildlife} & \multicolumn{2}{c|}{Vehicles}                                                                          & \multicolumn{2}{c|}{Medical}                                                                                         & \multicolumn{2}{c|}{Artworks} & \multicolumn{2}{c|}{Fashion}                                                                                            & Mean                        \\ \hline
&                        &                & \begin{tabular}[c]{@{}c@{}}CUB\\ (Acc)\end{tabular} & \begin{tabular}[c]{@{}c@{}}St-Dogs\\ (Acc) \end{tabular} 
           & \begin{tabular}[c]{@{}c@{}}iNat\\ (Acc)\end{tabular} & \begin{tabular}[c]{@{}c@{}}Cas-Leaf\\ (Acc)\end{tabular}

& \begin{tabular}[c]{@{}c@{}}St-Cars\\ (Acc) \end{tabular}& \begin{tabular}[c]{@{}c@{}}Air\\ (Acc)\end{tabular} 

& \begin{tabular}[c]{@{}c@{}}REF.-2 \\ (AUC)\end{tabular} & \begin{tabular}[c]{@{}c@{}}COV-19\\ (Acc)\end{tabular} 

& \begin{tabular}[c]{@{}c@{}}iMet\\ (F2)\end{tabular} & \begin{tabular}[c]{@{}c@{}}WikiArt\\ (Acc\%)\end{tabular} 

& \begin{tabular}[c]{@{}c@{}}Fash.Sty.\\ (Acc)\end{tabular} & \begin{tabular}[c]{@{}c@{}}DeepFash.\\ (Acc)\end{tabular}

&                             \\ \hline
\multicolumn{1}{c|}{BCN~\cite{dubey2018pairwise}}    &25M           & ResNet-50           & \cellcolor[HTML]{EFEFEF}87.7                        & \cellcolor[HTML]{EFEFEF}87.1  & \cellcolor[HTML]{EFEFEF}67.8 & \cellcolor[HTML]{EFEFEF}90.5 & \cellcolor[HTML]{EFEFEF}88.4   & \cellcolor[HTML]{EFEFEF}90.3                       & 77.7                                                    & 90.0                                                      & 54.8                                                          & 68.6                           & 78.7  & 79.7                    & 79.7                       \\
\multicolumn{1}{c|}{ACNet~\cite{ji2020attention}}   &48M           & ResNet-50           & \cellcolor[HTML]{EFEFEF}88.1                        & \cellcolor[HTML]{EFEFEF}90.7  & \cellcolor[HTML]{EFEFEF}70.3 & \cellcolor[HTML]{EFEFEF}89.0 & \cellcolor[HTML]{EFEFEF}91.2   & \cellcolor[HTML]{EFEFEF}92.4                       & 78.5                                                    & 90.7                                                      & 59.0                                                          & 73.8                               & 79.2  & 80.5                      & 81.2                        \\
\multicolumn{1}{c|}{PMG~\cite{du2020fine}}   &25M            & ResNet-50           & \cellcolor[HTML]{EFEFEF}89.6                        & \cellcolor[HTML]{EFEFEF}{\color[HTML]{000000} 91.7}  & \cellcolor[HTML]{EFEFEF}70.8     & \cellcolor[HTML]{EFEFEF}91.5 & \cellcolor[HTML]{EFEFEF}93.9   & \cellcolor[HTML]{EFEFEF}93.4 & 76.8                                                    & 88.0                                                      & 55.7                                                          & 69.2                               & 78.1  & 80.1                      & 80.6                        \\
\multicolumn{1}{c|}{API-NET~\cite{zhuang2020learning}}    &36M       & ResNet-50           & \cellcolor[HTML]{EFEFEF}87.7                        & \cellcolor[HTML]{EFEFEF}89.1   & \cellcolor[HTML]{EFEFEF}67.7     & \cellcolor[HTML]{EFEFEF}86.2 & \cellcolor[HTML]{EFEFEF}90.3   & \cellcolor[HTML]{EFEFEF}93.0                       & 77.9                                                    & 89.6                                                      & 56.5                                                          & 71.3                                     & 80.4  & 81.0                & 81.5                        \\
\multicolumn{1}{c|}{Cross-X~\cite{luo2019cross}}  &30M         & SENet-50            & \cellcolor[HTML]{EFEFEF}87.5                        & \cellcolor[HTML]{EFEFEF}92.2  & \cellcolor[HTML]{EFEFEF}71.1     & \cellcolor[HTML]{EFEFEF}90.8 & \cellcolor[HTML]{EFEFEF}94.6   & \cellcolor[HTML]{EFEFEF} 92.7                       & 79.5                                                    & 92.1                                                      & 60.5                                   & 73.9           & 80.9  & 81.7                   & 82.7                        \\
\multicolumn{1}{c|}{DCL~\cite{chen2019destruction}}     &28M          & ResNet-50           & \cellcolor[HTML]{EFEFEF}87.8                        & \cellcolor[HTML]{EFEFEF}88.8  & \cellcolor[HTML]{EFEFEF}67.4     & \cellcolor[HTML]{EFEFEF}87.5 & \cellcolor[HTML]{EFEFEF}91.2   & \cellcolor[HTML]{EFEFEF}88.6                    & 77.9                                                    & 90.2                                                      & 58.1                                                          & 70.9                                    & 78.6  & 79.2                 & 79.7                        \\
\multicolumn{1}{c|}{MGE~\cite{zhang2019learning}}    &28M            & ResNet-50           & \cellcolor[HTML]{EFEFEF}88.5                        & \cellcolor[HTML]{EFEFEF}90.7 & \cellcolor[HTML]{EFEFEF}69.1     & \cellcolor[HTML]{EFEFEF}89.8 & \cellcolor[HTML]{EFEFEF}93.7     & \cellcolor[HTML]{EFEFEF}89.2                   & 78.3                                                    & 93.0                                                      & 59.8                                                          & 74.5                                         & 81.6  & 81.2             & 81.5                        \\
\multicolumn{1}{c|}{Mix+~\cite{li2020attribute}}   &25M           & ResNet-50           & \cellcolor[HTML]{EFEFEF}88.4                        & \cellcolor[HTML]{EFEFEF}91.0  & \cellcolor[HTML]{EFEFEF}68.5     & \cellcolor[HTML]{EFEFEF}89.3 & \cellcolor[HTML]{EFEFEF}93.1 & \cellcolor[HTML]{EFEFEF}92.0                        & 79.0                                                    & 91.2                                                      & 56.0                                                          & 71.8                                     & 81.3  & 81.4                  & 81.7                        \\
\multicolumn{1}{c|}{TransFG\cite{he2022transfg}}  &86M   & ViT-B\_16           & \cellcolor[HTML]{EFEFEF}{91.7}                        & \cellcolor[HTML]{EFEFEF}{91.9} & \cellcolor[HTML]{EFEFEF}71.7     & \cellcolor[HTML]{EFEFEF}91.7 & \cellcolor[HTML]{EFEFEF}93.8 & \cellcolor[HTML]{EFEFEF}93.6 & 80.6                                                    & 91.8                                                      & 58.2                                                          & 74.5                                        & 82.0  & 81.7              &  83.4 \\
\multicolumn{1}{c|}{RAMS-Trans~\cite{hu2021rams}}  &86M & ViT-B\_16           & \cellcolor[HTML]{EFEFEF}{ 91.3} & \cellcolor[HTML]{EFEFEF}92.4     & \cellcolor[HTML]{EFEFEF}70.2     & \cellcolor[HTML]{EFEFEF}93.3 & \cellcolor[HTML]{EFEFEF}94.8 & \cellcolor[HTML]{EFEFEF}92.7                    & 79.9                                                    & 92.2                                                      & 59.8                                                          & 75.2                                     & 81.9  & 81.3                 & 83.0                        \\
\multicolumn{1}{c|}{DualCross~\cite{zhu2022dual}} &88M   & ViT-B\_16           & \cellcolor[HTML]{EFEFEF}{ 92.0} & \cellcolor[HTML]{EFEFEF}92.3  & \cellcolor[HTML]{EFEFEF}72.1     & \cellcolor[HTML]{EFEFEF}92.5 & \cellcolor[HTML]{EFEFEF}94.5 & \cellcolor[HTML]{EFEFEF}93.3                       & 80.8                                                    & 91.3                                                      & 59.5                                                          & 75.8                                        & 82.7  & 82.8              &  {83.8} \\ \hline
\multicolumn{1}{c|}{DualStage~\cite{bajwa2019two}}  &33M       & UNet + ResNet-50    & 86.1                                                & 87.9      & 63.2 & 83.3 & 86.0   & 84.7                                           & \cellcolor[HTML]{EFEFEF}80.3                            & 92.8                                                      & 57.7                                                          & -                                           & 76.2  & 74.3              & -                        \\
\multicolumn{1}{c|}{DENet~\cite{fu2018disc}}     &30M         & ResNet-50           & -                                                   & -      & - & - & -   & -                                               & \cellcolor[HTML]{EFEFEF}84.7                            & -                                                         & -                                                             & -                                                   & - & -     & -                           \\
\multicolumn{1}{c|}{FundTrans~\cite{fan2022detecting}}  &86M   & ViT-B\_16           & 90.1                                                & 91.8      & 65.3 & 86.5 & 88.1   & 86.9                                           & \cellcolor[HTML]{EFEFEF}{85.3}     & 93.5                                                      & 59.0                                                          & 74.1                                                  & 81.8 & 83.2   & 83.8                        \\
\multicolumn{1}{c|}{SatFormer~\cite{jiangsatformer}}    &92M      & ViT-B\_16           & 91.2                                                & 92.6    & 66.7 & 87.4 & 88.5   & 88.1                                               & \cellcolor[HTML]{EFEFEF}85.0                            & 94.3                                                      & 60.3                                                          & 74.7                                 & 82.0 & 80.1                    & 84.0                           \\
\multicolumn{1}{c|}{Convid-ViT~\cite{zhang2021transformer}} &88M   & Swin-B              & -                                                   & -        & - & - & -   & -                                              & -                                                       & \cellcolor[HTML]{EFEFEF}95.5                              & -                                                             & {\color[HTML]{3531FF} -}                      & - & -           & -                           \\
\multicolumn{1}{c|}{ConvidNet~\cite{ahmed2021convid}} &46M  & ConvidNet           & 83.2                                                & 81.0        & 61.5 & 80.2 & 83.1   & 81.5                                           & 78.4                                                    & \cellcolor[HTML]{EFEFEF}93.3                              & 55.2                                                          & 67.5                                       & 75.5 & 78.3              & 76.4                        \\
\multicolumn{1}{c|}{PSNet~\cite{tian2018psnet}}   &22M           & WRN                 & 86.7                                                & 87.9       & 64.1 & 83.6 & 85.8   & 85.2                                           & 80.5                                                    & \cellcolor[HTML]{EFEFEF}94.0                              & 56.7                                                          & 69.0                                    & 77.8 & 79.0                  & 79.1                        \\
\multicolumn{1}{c|}{Convid-Trans~\cite{shome2021covid}}   &86M    & ViT-B\_16           & 90.1                                                & 91.2    & 65.6 & 86.9 & 88.4   & 87.2                                              & 81.1                                                    & \cellcolor[HTML]{EFEFEF}95.0                              & 60.1                                                          & 74.6                                 & 80.5 & 80.4                      & 82.5                        \\
\multicolumn{1}{c|}{ResGANet~\cite{cheng2022resganet}}     &27M      & ResNet-50           & 82.9                                                & 90.0    & 64.7 & 83.8 & 84.3   & 87.2                                             & \cellcolor[HTML]{EFEFEF}82.9                            & \cellcolor[HTML]{EFEFEF}94.0                              & 58.4                                                          & 73.5                                     & 76.3 & 78.2                  & 78.9                        \\
\multicolumn{1}{c|}{SynMIC~\cite{zhang2019medical}}    &30M         & ResNet-50           & 87.1                                                & 91.6     & 67.3 & 86.1 & 87.5   & 84.8                                             & \cellcolor[HTML]{EFEFEF}83.6                            & \cellcolor[HTML]{EFEFEF}94.5                              & 59.8                                                          & 72.5                                    & 79.1 & 81.8                   & 81.9                        \\
\multicolumn{1}{c|}{SeATrans~\cite{wu2022seatrans}}  &96M    & UNet + ViT-B\_16    & 90.3                                                & 92.6        & 67.8 & 87.4 & 87.8   & 85.4                                         & \cellcolor[HTML]{EFEFEF}{87.6}                   & \cellcolor[HTML]{EFEFEF}{ 95.8}       & -                                                             & -                                                      & 76.1 & 77.5    & -                           \\ \hline
\multicolumn{1}{c|}{CLIP-Art~\cite{conde2021clip}}   &88M       & ViT-B\_32           & -                                                   & -          & - & - & -   & -                                          & -                                                       & -                                                         & \cellcolor[HTML]{EFEFEF}60.8                                  & -                                                & - & -          & -                           \\
\multicolumn{1}{c|}{MLMO~\cite{gao2021multi}}     &27M     & ResNet-50           & 84.8                                                & 87.1       & 60.1 & 77.5 & 79.6   & 75.4                                          & 77.8                                                    & 90.2                                                      & \cellcolor[HTML]{EFEFEF}60.3                                  & 75.2                                             & 75.1 & 76.3          & 79.2                        \\
\multicolumn{1}{c|}{GCNBoost~\cite{el2021gcnboost}}    &21M      & GCN       & 85.2                                                & 88.9      & 61.3 & 78.8 & 82.3   & 78.7                                            & 79.5                                                    & 92.6                                                      & \cellcolor[HTML]{EFEFEF}62.8                                  & 74.7                                            & 80.5 & 79.8           & 80.6                        \\
\multicolumn{1}{c|}{Pavel~\cite{pavel}}   &155M    & SENet*2 + PNasNet-5 & -                                                   & -        & - & - & -   & -                                            & -                                                       & -                                                         & \cellcolor[HTML]{EFEFEF}{67.2}                         & -                                                 & - & -         & -                           \\
\multicolumn{1}{c|}{DualPath~\cite{zhong2020fine}}     &29M     & ResNet-50           & 84.2                                                & 87.5    & 61.8 & 78.2 & 81.7   & 74.5                                             & 77.8                                                    & 91.3                                                      & 60.8                                                          & \cellcolor[HTML]{EFEFEF}81.5           & 72.3 & 72.6                    & 79.9                        \\
\multicolumn{1}{c|}{Two-Stage~\cite{sandoval2019two}}    &48M     & ResNet-50           & 86.7                                                & 90.4    & 64.2 & 84.4 & 86.1   & 84.9                                              & 83.9                                                    & 91.8                                                      & {62.7}                                                          & \cellcolor[HTML]{EFEFEF}81.3          & 78.1 & 76.6                      & 82.5                        \\
\multicolumn{1}{c|}{DeepArt~\cite{mao2017deepart}}     &45M      & Vgg16               & -                                                   & -      & - & - & -   & -                                               & -                                                       & -                                                         & 55.6                                                          & \cellcolor[HTML]{EFEFEF}77.2                      & - & -          & -                           \\
\multicolumn{1}{c|}{RASA~\cite{lecoutre2017recognizing}}       &23M       & ResNet-34           & 81.8                                                & 85.2  &58.3 & 75.1 & 76.0   & 72.8                                                  & 75.6                                                    & 87.0                                                      & 58.5                                                          & \cellcolor[HTML]{EFEFEF}78.9          & 71.1 & 75.8                     & 77.4                        \\
\multicolumn{1}{c|}{CrossLayer~\cite{chen2019recognizing}}   &42M     & Vgg16               & 81.1                                                & 84.7     &57.6 & 74.5 & 76.3   & 72.2                                            & 75.8                                                    & 86.8                                                      & {62.7}                                                          & \cellcolor[HTML]{EFEFEF}80.7      & 72.7 & 73.2                          & 77.7                        \\ \hline

\multicolumn{1}{c|}{EffNet~\cite{tan2019efficientnet}}     &66M       & EffNet-B7           & \cellcolor[HTML]{EFEFEF}91.0                        & \cellcolor[HTML]{EFEFEF}{ 92.0}  &\cellcolor[HTML]{EFEFEF}70.7 &\cellcolor[HTML]{EFEFEF}92.2 &\cellcolor[HTML]{EFEFEF}93.8   &\cellcolor[HTML]{EFEFEF}93.2 & \cellcolor[HTML]{EFEFEF}83.6                            & \cellcolor[HTML]{EFEFEF}{94.8}       & \cellcolor[HTML]{EFEFEF}61.0    & \cellcolor[HTML]{EFEFEF}81.2     & \cellcolor[HTML]{EFEFEF}82.8 & \cellcolor[HTML]{EFEFEF}82.6      & 85.4                      \\

\multicolumn{1}{c|}{ConvNeXtV2~\cite{woo2023convnext}}     &89M       & ConvNextV2-B           & \cellcolor[HTML]{EFEFEF}91.5                        & \cellcolor[HTML]{EFEFEF}{ 92.6}  &\cellcolor[HTML]{EFEFEF}73.5 &\cellcolor[HTML]{EFEFEF}93.4 &\cellcolor[HTML]{EFEFEF}94.7   &\cellcolor[HTML]{EFEFEF}94.0 & \cellcolor[HTML]{EFEFEF}84.3                            & \cellcolor[HTML]{EFEFEF}{96.0}       & \cellcolor[HTML]{EFEFEF}62.7 & \cellcolor[HTML]{EFEFEF}81.8 & \cellcolor[HTML]{EFEFEF}83.3 & \cellcolor[HTML]{EFEFEF}84.1  & 86.5               \\           

\multicolumn{1}{c|}{CvT~\cite{wu2021cvt}}    &32M      & CvT-21              & \cellcolor[HTML]{EFEFEF}90.6                        & \cellcolor[HTML]{EFEFEF}91.6   &\cellcolor[HTML]{EFEFEF}69.7 &\cellcolor[HTML]{EFEFEF}90.9 &\cellcolor[HTML]{EFEFEF}94.2   &\cellcolor[HTML]{EFEFEF}93.0                      & \cellcolor[HTML]{EFEFEF}82.1                            & \cellcolor[HTML]{EFEFEF}94.5                              & \cellcolor[HTML]{EFEFEF}61.5                                  & \cellcolor[HTML]{EFEFEF}{79.7}  & \cellcolor[HTML]{EFEFEF}82.5 & \cellcolor[HTML]{EFEFEF}83.8     & 84.8                        \\

\multicolumn{1}{c|}{DeiT~\cite{touvron2021training}}  &86M       & DeiT-B              & \cellcolor[HTML]{EFEFEF}91.1                        & \cellcolor[HTML]{EFEFEF}{ 92.1}  &\cellcolor[HTML]{EFEFEF}70.5 &\cellcolor[HTML]{EFEFEF}91.0 &\cellcolor[HTML]{EFEFEF}94.5   &\cellcolor[HTML]{EFEFEF}93.8 & \cellcolor[HTML]{EFEFEF}83.3                            & \cellcolor[HTML]{EFEFEF}{95.7}       & \cellcolor[HTML]{EFEFEF}60.7                                  & \cellcolor[HTML]{EFEFEF}{ 80.4}   & \cellcolor[HTML]{EFEFEF}83.7 & \cellcolor[HTML]{EFEFEF}82.8   & 85.3 \\

\multicolumn{1}{c|}{ViT~\cite{dosovitskiy2020image}}     &86M     & ViT-B\_16           & \cellcolor[HTML]{EFEFEF}90.3                        & \cellcolor[HTML]{EFEFEF}91.7  & \cellcolor[HTML]{EFEFEF}68.7     & \cellcolor[HTML]{EFEFEF}90.2 & \cellcolor[HTML]{EFEFEF}93.7 & \cellcolor[HTML]{EFEFEF}91.6                       & \cellcolor[HTML]{EFEFEF}81.8                            & \cellcolor[HTML]{EFEFEF}93.8                              & \cellcolor[HTML]{EFEFEF}60.4                                  & \cellcolor[HTML]{EFEFEF}78.8              & \cellcolor[HTML]{EFEFEF}81.1 & \cellcolor[HTML]{EFEFEF}82.6               & 82.8                        \\ \hline
                     \\ 
\multicolumn{1}{c|}{Pro-NeXt-B/8}  &86M   & Pro-NeXt-B        & \textbf{93.3} &  \textbf{94.5}  
& \textbf{75.8}     & \textbf{94.7} & \textbf{96.3} & \textbf{95.6} &  \textbf{87.8}     &  \textbf{96.7}       &  \textbf{67.9}       & \textbf{82.7}  & \textbf{85.3} & \textbf{85.6}    & \textbf{88.0}                        \\ 
\hline
\bottomrule
\end{tabular}}
\vspace{3pt}
\end{table*}


\begin{figure*}[hbt!]
\centering
\includegraphics[width=0.85\linewidth]{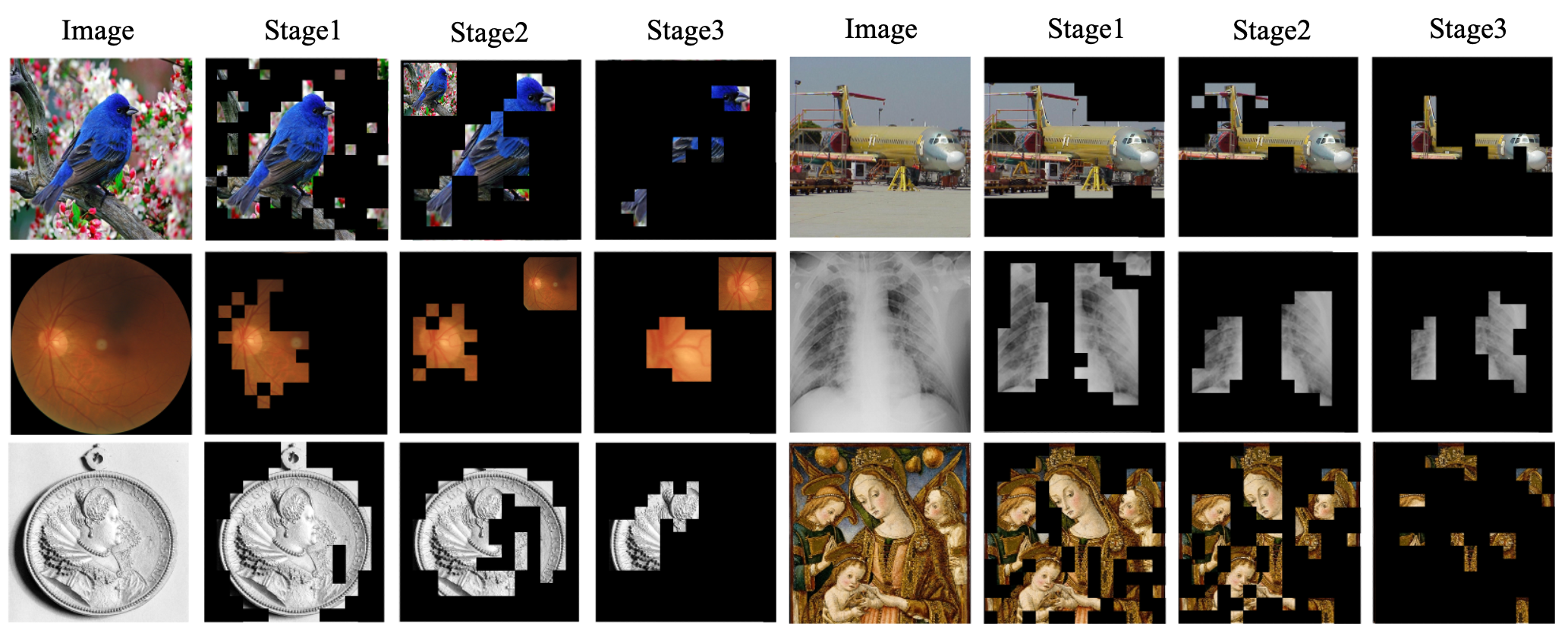}
\caption{The visualized results of Shift-Parser mask in each stage (mapping back to the raw image). The focal parts are kept and the context parts are masked. For the small focal parts, we zoom in the region on the up-left/right corners for clarity. From top to down are wildlife/vehicles, medical image, and artwork classification respectively.}
\label{fig:mask_vis}
\vspace{-15pt}
\end{figure*}


\subsection{Generalization Ability}
Pro-NeXt distinguishes itself from task-specific models with its strong generalization across diverse domains, as shown in Table~\ref{tab:mainres}. Unlike FGVC methods that often overfit to wildlife and vehicle datasets, and perform a noticeable decline applied to other professional domains. Pro-NeXt excels across professional fields. In medical imaging, Pro-NeXt achieves top performance on COVID-19 and REFUGE2 datasets, surpassing even segmentation required method, SeATrans. For art and fashion datasets, Pro-NeXt also ranks first on most benchmarks, which even outperforms Pavel~\cite{pavel} that uses treble the parameters.

We also compare our method with the general vision classification architectures. We take ViT based models ~\cite{dosovitskiy2020image, touvron2021training}, convolution based models~\cite{tan2019efficientnet, woo2023convnext}, and also hybird model~\cite{wu2021cvt} for the comparison. We can see these methods often show stronger generalization ability, but perform worse than SOTA on the specific tasks. In contrast, Pro-NeXt performs well on all three tasks, and shows the best generalization ability. Comparing the mean score with these methods, Pro-NeXt outperforms SOTA DeiT by 2\% with half of the parameters.

\subsection{Scalability}
\begin{figure}[ht]
    \begin{minipage}{0.5\linewidth}
            \centering
            \includegraphics[scale=0.50]{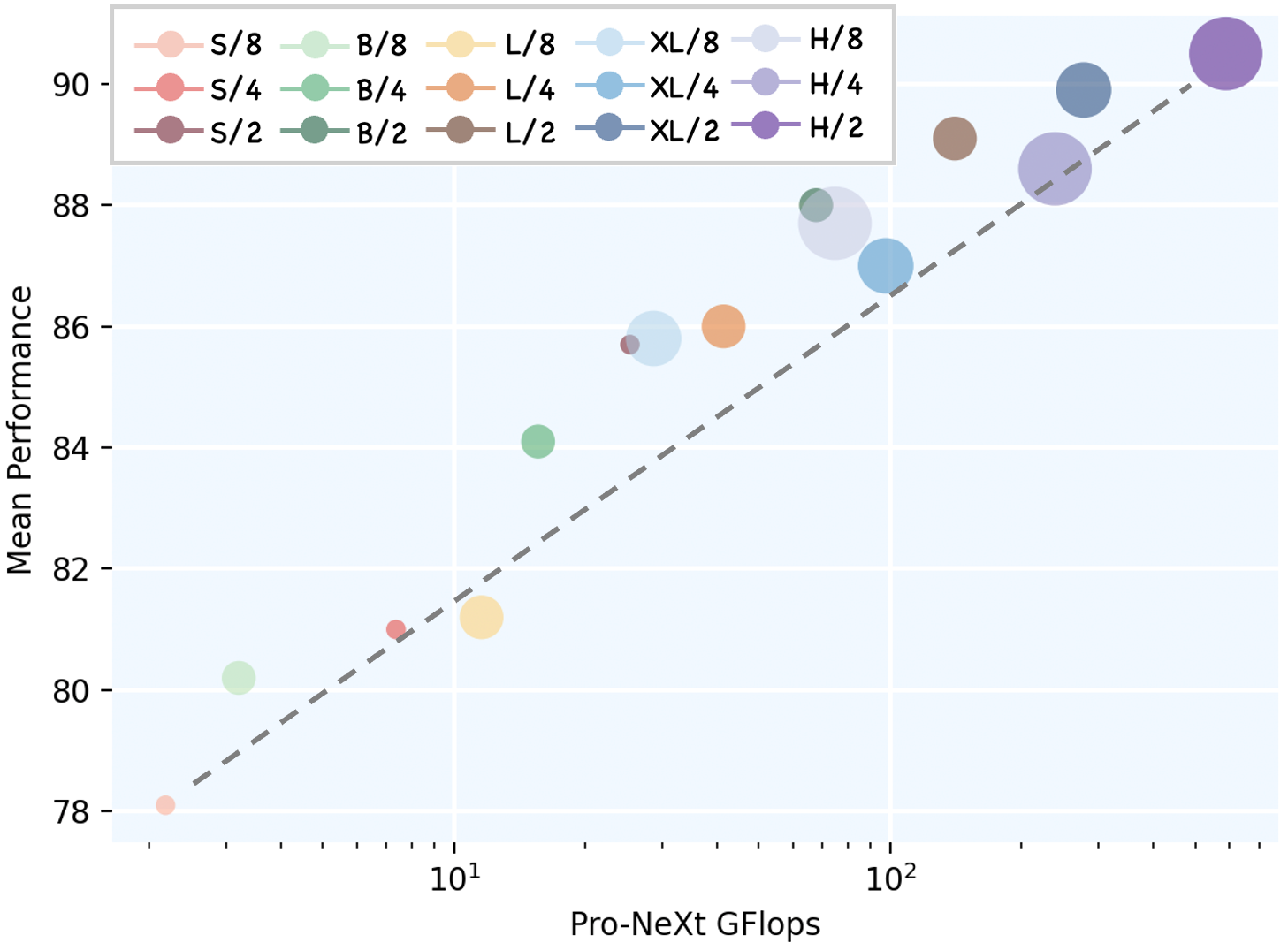}
            \vspace{-5pt}
             \caption{Pro-NeXt Gflops are strongly correlated with performance. We plot the Gflops of each of our Pro-NeXt models and each model’s mean performance after 200K training steps.}
            \label{fig:scale}
    \end{minipage}\,
    \hfill
\begin{minipage}{0.5\linewidth}
        \centering
        \captionof{table}{Scalability of the model. Pro-NeXt shows strong model scaling behavior, with consistently improved performance across all model sizes. }
        \vspace{10pt}
        \resizebox{0.95\textwidth}{!}{
        \begin{tabular}{ccccc}
        \hline
        model      & param & FLOPs & \begin{tabular}[c]{@{}c@{}}throughput\\ (image/s)\end{tabular} & \begin{tabular}[c]{@{}c@{}}performance\\ (Ave)\end{tabular} \\ \hline
        RegNetY16G                                                       &  84M     &  48.1G     &    78.7                                                            &   83.0                                                     \\
        EffNet-B7                                                        &  66M     &  34.0G     &       61.1                                                         &    85.4                                                    \\
        EffNetV2-L                                                       &  120M     &  51.0G     &     86.9                                                           &  86.9                                                      \\ \hline
        DeiT-S                                                           &  22M     &  14.7G     &    338.2                                                            &   83.2                                                    \\
        DeiT-B                                                           &  86M     &  53.3G     &    76.5                                                            & 85.1                                                       \\ 
        CvT-S                                                           &  20M     &  16.1G     &    305.6                                                            & 83.1                                                       \\ 
        CvT-B                                                           &  32M     &  28.8G    &    226.5                                                            & 84.8                                                       \\
        CvT-H                                                          &  277M     &  185.0G     &    38.7                                                            & 87.1                                                       \\ \hline
        ConvNeXt-V2-S                                                           & 29M    &    24.2G   &     268.5                                                           & 85.5                                                       \\
        \cellcolor[HTML]{EFEFEF}Pro-NeXt-S/2                                                       &  \cellcolor[HTML]{EFEFEF}29M     &  \cellcolor[HTML]{EFEFEF}25.3G     &   \cellcolor[HTML]{EFEFEF}263.9                                                             &  \cellcolor[HTML]{EFEFEF}\textbf{85.7}                                                      \\
        ConvNeXt-V2-B                                                           & 89M    &   61.9G    &   87.0                                                             &  86.5                                                      \\
        \cellcolor[HTML]{EFEFEF}Pro-NeXt-B/2                                                        & \cellcolor[HTML]{EFEFEF}86M    &    \cellcolor[HTML]{EFEFEF}67.6G   &   \cellcolor[HTML]{EFEFEF}84.4                                                            &  \cellcolor[HTML]{EFEFEF}\textbf{88.0}                                                      \\
        ConvNeXt-V2-L                                                           & 198M    &   138.4G    &   68.5  & 87.4 \\
        \cellcolor[HTML]{EFEFEF}Pro-NeXt-L/2                                                        & \cellcolor[HTML]{EFEFEF}182M   &   \cellcolor[HTML]{EFEFEF}140.6G    & \cellcolor[HTML]{EFEFEF} \cellcolor[HTML]{EFEFEF}61.9                                                            &   \cellcolor[HTML]{EFEFEF}\textbf{89.1}                                                     \\
        ConvNeXt-V2-XL                                                           & 350M   &  227.6G     &  28.5                                                              &  87.9                                                      \\
        \cellcolor[HTML]{EFEFEF}Pro-NeXt-XL/2                                                        & \cellcolor[HTML]{EFEFEF}321M   &   \cellcolor[HTML]{EFEFEF}213.8G    &  \cellcolor[HTML]{EFEFEF}37.4                                                              &  \cellcolor[HTML]{EFEFEF}\textbf{89.9}\\ 
        ConvNeXt-V2-H                                                           & 659M   &  587.8G     &  6.7                                                              &  88.2                                                      \\
        \cellcolor[HTML]{EFEFEF}Pro-NeXt-H/2                                                        & \cellcolor[HTML]{EFEFEF}634M   &  \cellcolor[HTML]{EFEFEF} 577.1G    & \cellcolor[HTML]{EFEFEF} 7.4                                                              &  \cellcolor[HTML]{EFEFEF}\textbf{90.5}\\ 
        \hline                                                     
        \end{tabular}}\label{table:scale}
    \end{minipage}
\vspace{-15pt}
\end{figure}

We explore the Pro-NeXt design space and investigate the scaling properties of our model class. A scaling experiment was conducted with 15 Pro-NeXt models, each varying in model configurations (S, B, L, XL, H) and patch sizes (8, 4, 2). We find that scaling model Gflops is the key to improved performance. In Fig. \ref{fig:scale}, we plot the mean performance on all test benchmarks against model Gflops. The overarching observation is that an increase in model size coupled with a reduction in patch size consistently yields improved model performance. The results also indicate that different Pro-NeXt configurations achieve similar performance when their total Gflops are comparable (e.g., S/2 and XL/8). This robust correlation observed between model Gflops and performance implies that increased model compute is pivotal for enhancing Pro-NeXt models.

Furthermore, we compare Pro-NeXt with previous scalable models with different model parameters. The results are presented in Table \ref{table:scale}. In comparison with convolution-based architectures, Pro-NeXt-B/2 outperforms 120M EffNetV2-L by 1.1\% with a smaller 86M. Compared to transformer-based architectures, Pro-NeXt-L/2 surpasses 277M CvT-H by 2\% with a smaller 182M. Notably, when compared to ConvNext-V2, Pro-NeXt shows greater improvement with increasing model size: the small-size Pro-NeXt-S/2 outperforms ConvNext-V2-S by 0.3\%, while the huge-size Pro-NeXt-H/2 outperforms the corresponding ConvNext-V2-H by 2.3\%, indicating the stronger scalability of Pro-NeXt.

\begin{figure}[h]
\centering
\includegraphics[width=0.85\linewidth]{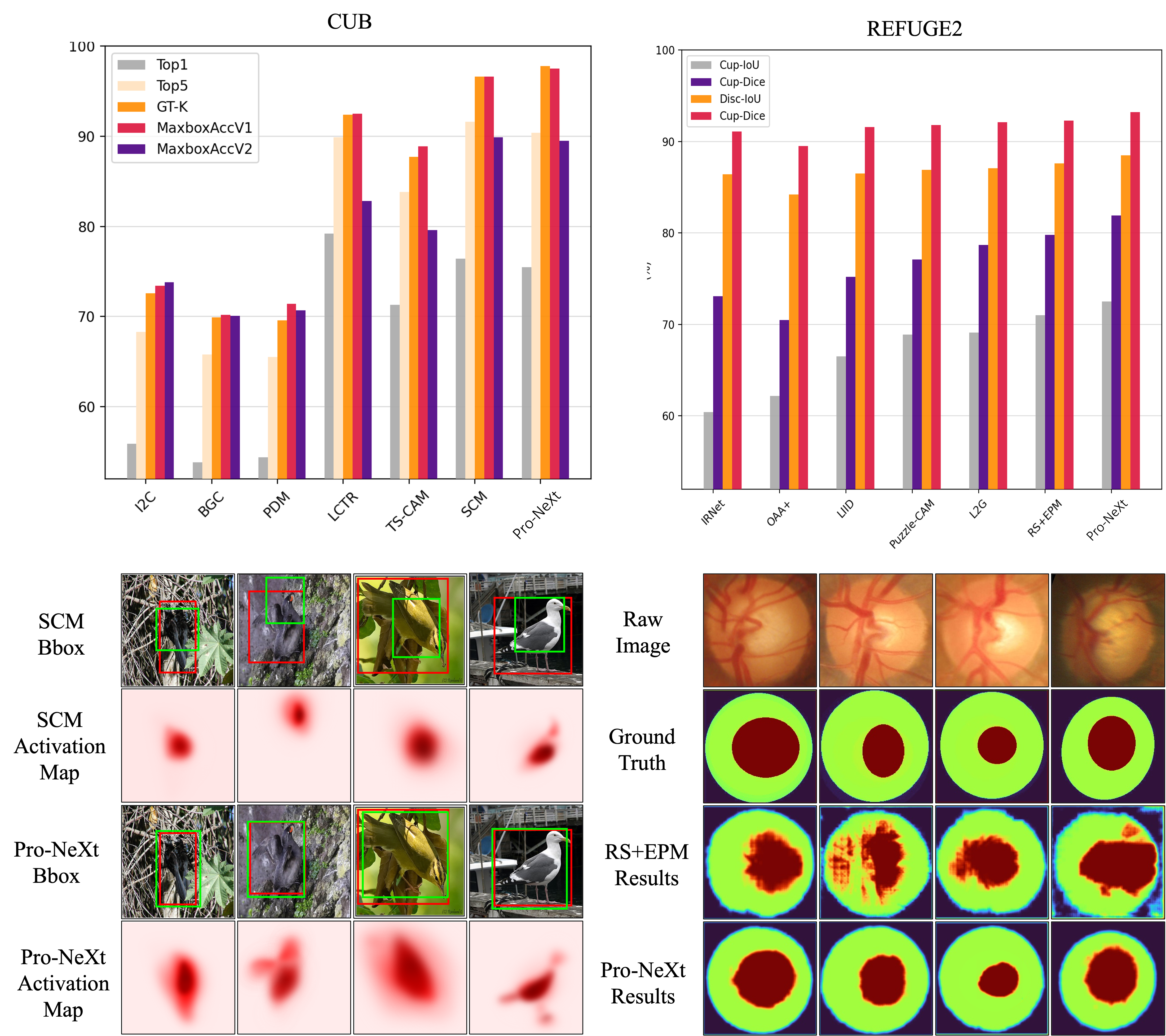}
\caption{Quantitative and qualitative comparison of explainability on CUB and REFUGE2, respectively. On CUB, we show activation score map and bounding box (Bbox) of SCM for comparison. The ground-truth and predicted Bboxes are shown as {\color[HTML]{FE0000}red} and {\color[HTML]{32CD32}green} respectively. On REFUGE2, we show activated optic disc ({\color[HTML]{00FF00}green}) and cup ({\color[HTML]{800020}brown}) segmentation of RS+EPM for the comparison. 
}
\label{fig:increment}
\end{figure}


\subsection{Explanability}\label{sec:exp_prior}
During the analysis of Pro-NeXt, we have observed from that Pro-NeXt can produce strongly explainable parser mapping in its inference. We show some visualized results of the intermediate Shift-Parser in Fig. \ref{fig:mask_vis}. The masks are mapped back to the raw image space for analysis. We observe that the inference process of Pro-NeXt is quite simular with that of the human domain experts.

We can see in the case of glaucoma, Pro-NeXt prioritizes the optic disc in Stage2 and the optic cup in Stage3, so that the fusion of the focal embedding and the final Complex Impression can effectively capture the relationship between the optic disc and cup. The vertical optic disc and cup ratio (vCDR) is exactly a indication to diagnose glaucoma clinically. In the case of painting, Pro-NeXt individually models the background, the figure faces/gestures, and ornaments in the three stages. The serene facial expressions, detailed drapery, and meticulous use of gold leaf in the painting serve as strong evidence of its origins in the Italian Early Renaissance.

It opens a new window that Pro-NeXt can naturally generate the explainable visualized results without using Class-Activation Map (CAM) based techniques. We then further evaluate its explanability using the intermediate parser map $m$ to produce object segmentation or localization results over Pro-NeXt. The evaluation is conducted on bird species recognition and glaucoma prediction. We show the quantitative results and visual comparison in Fig.~\ref{fig:increment}. We produce the bird location and activation from the first-stage map $m^{1}$ on CUB dataset, and produce the optic disc and cup segmentation prediction from $m^{2}$ and $m^{3}$ respectively on REFUGE2 dataset (The detailed produce procedure is given in the supplementary). On CUB dataset, we compare it against the mainstream explainable classification methods, including l2C~\cite{zhang2020inter}, BGC~\cite{kim2022bridging}, PDM~\cite{meng2022diverse}, LCTR~\cite{chen2022lctr}, TS-CAM~\cite{gao2021ts}, and SCM~\cite{bai2022weakly}. We also evaluated their quantitative performance by the commonly used metrics, including GT-Known (GT-K), Top1/Top5 Localization Accuracy, and more strict ones like MaxboxAccV1 and MaxboxAccV2~\cite{choe2020evaluating}. On REGUGE2 dataset, we compare it against the explainable methods with segmentation productions, including IRNet~\cite{ahn2019weakly}, OAA++~\cite{jiang2019integral}, LIID~\cite{liu2020leveraging}, Puzzle-CAM~\cite{jo2021puzzle}, L2G~\cite{jiang2022l2g}, and RS+EPM~\cite{jo2022recurseed} through IoU and Dice.

We can see on the CUB dataset, Pro-NeXt outperforms SOTA SCM with 1.2\% over GT-K and 1.1\% over MaxboxAccV1. The visualized results show that Pro-NeXt predicts more sophisticated activation maps, and thus produces more accurate bounding boxes. On REFUGE2 dataset, Pro-NeXt outperforms SOTA RS+EPM with 2.5\% over IoU and 2.1\% over Dice. Compared with RS+EPM over the visualized results, Pro-NeXt predicts neater and more reasonable segmentation masks, especially on the ambiguous optic cup.

\begin{table}[h]
\centering
\caption{Ablation study over Gaze-Shift and Shift-Parser. SA denotes Spatial Attention. Cond. Sine denotes Conditional Sine.}
\resizebox{0.85\textwidth}{!}{
\begin{tabular}{c|cc|cc|ccc}
\toprule
\hline
Baseline & \multicolumn{2}{c|}{Gaze-Shift}                                                                                & \multicolumn{2}{c|}{Shift-Parser}                                            & Wildlife  & Med & Art \\ \hline
         & \begin{tabular}[c]{@{}c@{}}Focal-Context\\ Separation\end{tabular} & \begin{tabular}[c]{@{}c@{}}Context\\ Impression\end{tabular} & \begin{tabular}[c]{@{}c@{}}Cond. Sine\\ Activation\end{tabular} & \begin{tabular}[c]{@{}c@{}}Band-Pass\\ Filter\end{tabular} & Ave     &  Ave     & Ave     \\ \hline
\checkmark     &                                                           &                                                      &                                                               &           & 83.7 & 84.3  & 68.7 \\
\checkmark     & \checkmark                                                      &                                                      &     \multicolumn{2}{c|}{SA}           & 81.2 & 87.6  & 70.7 \\
\checkmark     & \checkmark                                                      & \checkmark                                                 &   \multicolumn{2}{c|}{SA}                                                                       & 85.1 & 89.7  & 72.3 \\
\checkmark     & \checkmark                                                      & \checkmark                                                 &   \multicolumn{2}{c|}{NeRF}                                                                       & 88.1 & 90.0  & 72.9 \\
\checkmark     & \checkmark                                                      & \checkmark                                                 & \checkmark                                                          &           & 88.4 & 91.6  & 74.3 \\
\checkmark     & \checkmark                                                      & \checkmark                                                 & \checkmark                                                          & \checkmark      & \textbf{89.6} & \textbf{92.3}  & \textbf{75.4} \\ \hline \bottomrule
\end{tabular}}\label{tab:ablation}
\vspace{-20pt}
\end{table}


\subsection{Ablation Study}
We provide quantitative ablation study results in Table \ref{tab:ablation}. The results are evaluated by the average performance of the three domains. We begin with a ResNet baseline and sequentially add the proposed modules over the backbone, observing gradual improvement in model performance. We first adopt Gaze-Shift with only Focal-Context Separation, using basic Spatial Attention (SA)\cite{woo2018cbam} as the filter. We can see that Gaze-Shift significantly improves the performance on tasks where certain parts are discriminative (Wildlife and Medical). However, we surprisingly find that SA leads to worse performance than baseline on FGVC, as verified by many others in the community\cite{WinNT}. This may be because soft attention assigns low weights to context in each stage, thus failing to capture the interaction between the focal and context parts that is crucial for recognition. In comparison, Gaze-Shift with feature separation can learn this interaction explicitly, overcoming the limitations of local-part-based classification. This outcome demonstrates the necessity of the binary map in Gaze-Shift.

By then applying context impression based on channel attention, we observe a consistent improvement in performance, indicating that global attention is more effective for modeling contextual information. After that, we replace the SA module in our framework with Shift-Parser, which leads to a significant performance boost. Compared to the baseline NeRF, Shift-Parser achieves significant improvements of 1.5\%, 2.3\%, and 2.5\% on three domains, as shown in Table \ref{tab:ablation}, demonstrating the effectiveness of our proposed conditional sine activation and band-pass filter. More detailed ablation study is given in our supplementary.

\subsection{Discussion}
\subsubsection{Gap between Professional and General Recognition}
We also validate the performance of Pro-NeXt on general recognition benchmarks, like ImageNet, ImageNet ReaL \cite{beyer2020we}, and ImageNet V2 \cite{recht2019imagenet}. The results are shown in Table \ref{tab:general}. As shown in the table, Pro-NeXt-H achieves comparable performance with strong ConvNeXt-V2 competitor on ImageNet benchmarks. Notably, Pro-NeXt exhibits stronger scalability, outperforming ConvNeXt-V2 as the scale increases. Additionally, our observations indicate that ConvNeXt-V2 does not perform as effectively on Pro-NeXt tasks. It again shows a domain gap exists between general and professional recognition.

\setcounter{footnote}{0}
Better performance on general image classification does not necessarily translate to better performance on professional recognition tasks. These tasks have unique features that have not been fully exploited by general image classification methods. For instance, CBAM \cite{woo2018cbam}, widely used in general image classification, demonstrates subpar performance in FGVC tasks, as experimentally verified in our ablation study, and also verified by many other users in the community\footnote{https://github.com/Jongchan/attention-module/issues/24}. This may because the commonly used attention mechanism in general recognition would assign low weights to context in each stage, thus failing to capture the interaction between the focal and context parts that is crucial for recognition. In comparison, we design Pro-NeXt to learn this interaction explicitly to overcome the limitations of local-part-based classification.

\subsubsection{Machine Visual Hierarchy Comparing with Human}
\begin{figure}[ht]
    \begin{minipage}{0.5\linewidth}
    \centering
    \captionof{table}{Accuracy on ImageNet Benchmarks comparing with general image classification methods.}
    \resizebox{\textwidth}{!}{
    \begin{tabular}{c|cccc}
\hline
 Model & ImageNet  &ImageNet ReaL &ImageNet V2& Ave on FG\\ \hline
ResNet50 & 76.2      & 82.4  &63.9               & 78.8                        \\
Mixer-B & 61.7  & 73.9 &53.4 & 69.1  \\ 
ViT-B & 80.1 &85.5                & 68.1  &82.8\\  \hline
ConvNeXt-V2-B                                                           & \textbf{84.6}    &   \textbf{88.1}    &   \textbf{72.1}                                                             &  86.5                                                      \\
\cellcolor[HTML]{EFEFEF}Pro-NeXt-B                                                        & \cellcolor[HTML]{EFEFEF}84.2    &    \cellcolor[HTML]{EFEFEF}87.6   &   \cellcolor[HTML]{EFEFEF}71.6                                                            &  \cellcolor[HTML]{EFEFEF}\textbf{88.0}                                                      \\
ConvNeXt-V2-L                                                           & \textbf{85.6}    &   \textbf{88.9}    &   \textbf{72.6}  & 87.4 \\

\cellcolor[HTML]{EFEFEF}Pro-NeXt-L                                                        & \cellcolor[HTML]{EFEFEF}85.4   &   \cellcolor[HTML]{EFEFEF}88.6    & \cellcolor[HTML]{EFEFEF} \cellcolor[HTML]{EFEFEF}72.3                                                           &   \cellcolor[HTML]{EFEFEF}\textbf{89.1}                                                     \\
ConvNeXt-V2-H                                                           & \textbf{85.8}   &  \textbf{89.0}     &  72.8                                                              &  88.2                                                      \\
\cellcolor[HTML]{EFEFEF}Pro-NeXt-H                                                        & \cellcolor[HTML]{EFEFEF} \textbf{85.8}   &  \cellcolor[HTML]{EFEFEF} 88.9    & \cellcolor[HTML]{EFEFEF} \textbf{73.0}                                                              &  \cellcolor[HTML]{EFEFEF}\textbf{90.5}\\ 
\hline
    \end{tabular}}
    \label{tab:general}
    \end{minipage}\,
    \hfill
\begin{minipage}{0.5\linewidth}
        \centering
    \includegraphics[width=0.85\textwidth]{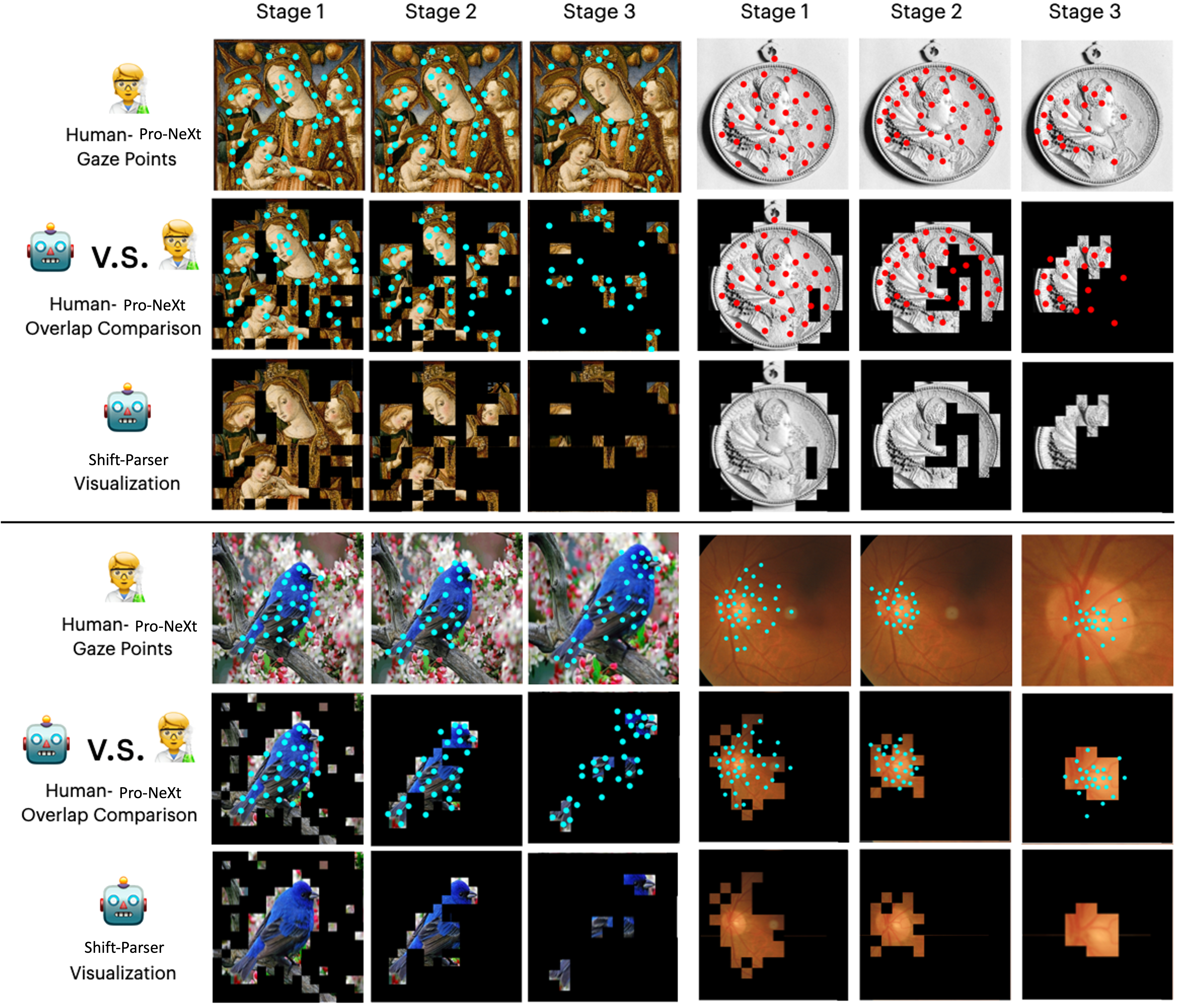}
    \caption{The comparison of human-experts and Pro-NeXt Shift-Parser visualized results. Blue or red dots denote experts' gaze positions.}
    \label{fig:my_label}
    \vspace{-15pt}
    \end{minipage}
\vspace{-15pt}
\end{figure}

Given that our model is inspired by the human visual perception, we undertook an experiment to evaluate the correspondence between our constructed artificial visual hierarchy and that of human specialists. We gathered gaze data from a group of four experts, including a seasoned ornithologist, a qualified ophthalmologist, and two professors specializing in historical archaeology. This was accomplished using the Tobii Pro Spectrum and implementing a sequential open-set clustering technique as detailed in \cite{gaze-track} for a stage-by-stage comparative analysis. The intersections between the regions focused on by the experts' gazes and the areas activated by our Shift-Parser were examined, with the findings illustrated in Fig. \ref{fig:my_label}. The results reveal that the recognition process employed by Pro-NeXt is reflective of the expert human process, suggesting a potential congruence between Pro-NeXt's operational mechanism and human perceptual abilities. This congruence likely underpins the robust explanatory power of the Pro-NeXt models.

\section{Conclusion}
In summary, this work introduces Pro-NeXt, a model poised to revolutionize Professional Visual Recognition by transcending the limitations of task-specific models through its generic, scalable and explainable design. Pro-NeXt's biologically-inspired architecture ensures strong generalizability and performance across varied fields such as fashion, medicine, and art, outshining previous models on multiple datasets. Our results underscore Pro-NeXt's exceptional scalability and explainability, aspects that have been largely overlooked in prior research within this field. These advantageous qualities of Pro-NeXt are poised for application on more computationally intensive resources and larger, more diverse datasets to achieve broader impact.

\bibliographystyle{splncs04}
\bibliography{main.bib}
\end{document}